\pdfoutput=1

\documentclass[11pt]{article}

\usepackage[]{acl}

\usepackage{times}
\usepackage{latexsym}
\usepackage{graphicx}
\usepackage[T1]{fontenc}
\usepackage[utf8]{inputenc}
\usepackage{microtype}
\usepackage{lipsum} 
\usepackage[]{mdframed}

\newcommand{\merel}[1]{\textcolor{black}{#1}}
\newcommand{\framedtext}[1]{
\par
\noindent\fbox{
    \parbox{\dimexpr\linewidth-2\fboxsep-2\fboxrule}{#1}
}
}

\title{Prompting Implicit Discourse Relation Annotation}

\author{Frances Yung\textsuperscript{1} \,
Mansoor Ahmad\textsuperscript{1} \,
Merel Scholman\textsuperscript{1,2} \,
Vera Demberg\textsuperscript{2}\\
\textsuperscript{1}Saarland University, Saarbr\"ucken, Germany\\
\textsuperscript{2}Utrecht University, Utrecht, Netherlands\\
\texttt{\{frances, mahmad, m.c.j.scholman, vera\}@coli.uni-saarland.de}}

\begin{document}
\maketitle
\begin{abstract}

Pre-trained large language models, such as ChatGPT, archive outstanding performance in various reasoning tasks without supervised training and were found to have outperformed crowdsourcing workers. Nonetheless, ChatGPT's performance in the task of implicit discourse relation classification, prompted by a standard multiple-choice question, is still far from satisfactory and considerably inferior to state-of-the-art supervised approaches.  This work investigates several proven prompting techniques to improve ChatGPT's recognition of discourse relations. In particular, we experimented with breaking down the classification task that involves numerous abstract labels into smaller subtasks.  Nonetheless, experiment results show that the inference accuracy hardly changes even with sophisticated prompt engineering, suggesting that implicit discourse relation classification is not yet resolvable under zero-shot or few-shot settings.

\end{abstract}

\section{Introduction}

Pre-trained language models have demonstrated superior performance in various NLP tasks for years, and recently prompt-tuning instead of fine-tuning has become the dominant framework to make efficient use of large language models (LLMs).  LLMs such as ChatGPT have demonstrated human-level performance in various reasoning tasks under zero-shot or few-shot settings using natural language prompts as inputs \citep[see e.g.,][]{openai2023gpt4,bang2023multitask}. This has led to a wave of research in prompt engineering to elicit the prediction potential of LLMs \citep[such as][]{wei2022chain,kojima2022large}.

In order to create metadata for textual analysis or to train models for specific NLP tasks, researchers have been relying on the annotation performed by trained annotators or crowdsourced workers. Recently, ChatGPT was shown to outperform crowdsourced workers in annotating political topics, affiliation, and policy frames \cite{gilardi2023chatgpt,tornberg2023chatgpt}.
However, it is not yet clear whether a similar prompting approach can also be successful for classifying what discourse relation holds between two text spans.
Discourse relations (DRs) are semantic-pragmatic links between clauses and sentences.  They can be explicitly marked by discourse connectives (DCs), such as \textit{however} and \textit{in addition}, or they can be inferred from the text without relying on a specific marker -- such cases are referred to as implicit relations.
For example, there is a \texttt{causal} relation between the following sentences: \textit{Mary lost her keys. Therefore, she could not enter her office.}, and the same relation can still be inferred without the DC \textit{therefore}. 

Discourse relation analysis is useful for various downstream tasks, such as summarization \cite{xu2020discourse,dong2021discourse} and relation extraction \cite{tang2021discourse}, and discourse-annotated data serves as the basis of various linguistic research \citep[e.g.][]{sanders2010discourse}.
However, classifying implicit DRs involves cognitive processing that is difficult even for humans in different languages \cite{oza2009experiments,zhou2012pdtb,polakova2013introducing,zeyrek2020ted,hoek2021there} and poses a challenge for NLP \citep[e.g., $64.58\%$ accuracy and $49.03\%$ F1 on PDTB 2.0 in][]{chan-etal-2023-discoprompt}, even with powerful LLMs.

\citet{chan2023chatgpt} evaluated ChatGPT's ability to infer implicit DRs. They used a multiple-choice prompt that lists all options of DR labels and included in-context samples.  They found that pairing the DR options with typical DCs improves the performance.  
However, even then, accuracy is still far behind (e.g., $24.54\%$ accuracy and $16.20\%$ F1 on PDTB 2.0) the performance of state-of-the-art supervised models. There could be several reasons for this: the chosen prompts might not be optimal, the LLM may not be able to deal well with a 14-way classification, or it may fail to build good representations of discourse relations.

The current work tests whether alternative prompt designs, using formulations that have been shown to work well in crowdsourced annotation tasks with untrained humans, can produce more accurate implicit DR annotation.  First, we replicate the methodology used by \citet{chan2023chatgpt} with the latest \textit{gpt-4} model and analyse the strengths and weaknesses of the predictions.  We then explore the benefits of breaking down the multi-way classification task into individual prompts. We hypothesize that this might be beneficial because a) it reduces a choice between 14 options per prompt into multiple prompts that each contain a binary choice and b)  because implicit DRs are inherently ambiguous and multiple interpretations are often possible depending on the reader's perspective and context \cite{rohde2016,scholman2022DiscoGeM}. 

We however found that sophisticated prompt strategies did not improve the performance of ChatGPT's inference of implicit DRs and the accuracy still largely lags behind the state-of-the-art supervised models.  This suggests that the implicit DR recognition task is not yet solvable under a zero-shot or few-shot setting.  

\section{Related Work}

\subsection{Lexicalized DR annotation}

DCs are natural language signals for DRs, and have been used by annotators to classify originally implicit DRs.  For example, the DC \textit{nevertheless} can be inserted between the two text spans that make up a DR (known as the \textit{Arg1} and \textit{Arg2}) to indicate a \texttt{concession} relation. 
This approach has been used in the creation of manually annotated resources including the Penn Discourse Treebank \citep[PDTB,][]{prasad-etal-2008-penn,AB2/SUU9CB_2019} and DiscoGeM \cite{scholman2022DiscoGeM}. 

A complicating factor in the annotation of DRs is the fine-grainedness between DR types.
Compared with typical text classification tasks such as entailment and stance detection, DR classification involves a considerably larger range of labels, making the task particularly challenging.  For example, the DR framework PDTB \cite{prasad-etal-2008-penn,AB2/SUU9CB_2019} defines
$36$ DR sense labels arranged in a three-level hierarchy, which can be lexicalized by one of $184$ connectives. Previous annotation efforts for implicit relations report inter-annotator agreement scores of
$\kappa$=$0.58$ \cite{hoek2021there} and $\kappa$=$0.47$ \citep{zikanova2019explicit} between expert annotators, and $\kappa$=$0.55$ between aggregated crowdsourced and expert annotation \cite{scholman2022DiscoGeM}.

Untrained human annotators also struggle when given a large choice of different relations or connectives. 
The DiscoGeM corpus was therefore annotated using a ``two-step DC insertion method'' \citep{yung-etal-2019-crowdsourcing}, where the crowd annotators first freely type a linking phrase that represents the relation between two consecutive sentences and then disambiguate this intuitive choice by selecting from a list of DC options dynamically generated based on the first DC. This approach lexicalizes DRs into natural language for untrained workers who are not familiar with DR labels. 

While general-purpose generative LLMs such as ChatGPT might have seen discourse relation labels as part of their training data, they could nevertheless profit from a connective-based annotation approach, as their exposure to connectives in natural language is much higher.
We thus evaluate the two-step DC insertion method on generative LLMs.

\subsection{Supervised models for DR recognition}

Given that explicit as well as originally implicit relations are annotated with a connective in the PDTB,
earlier work has explored the use of both explicit and annotated DCs for implicit DR recognition  \citep[e.g.,][]{marcu-echihabi-2002-unsupervised,sporleder2008using,xu2012connective,rutherford-xue-2015-improving,ji-etal-2015-closing}.  In combination with modern LLMs, DC prediction was found to be an effective sub-task for which the LLMs are fine-tuned to identify implicit DRs \cite{shi-demberg-2019-learning,kishimoto-etal-2020-adapting,jiang2021generating,kurfali-ostling-2021-lets,liu-strube-2023-annotation}. 

Recent improvements in supervised implicit DR recognition use prompt tuning based on pre-trained LLMs.  Typically, the prompt includes a template where the connective between two input sentences is to be predicted \cite{xiang-etal-2022-connprompt,zhao-etal-2023-infusing, wu-etal-2023-connective}. Other works also evaluated left-to-right generational prompts such as \textit{the connective between Arg1 and Arg2 is ...} \cite{zhou-etal-2022-prompt-based} or \textit{the sense between Arg1 and Arg2 is ...} \cite{xiang-etal-2023-teprompt}.  A list of less ambiguous connectives were selected to verbalize each DR sense label. In addition to DCs, hierarchical information of the sense labels was also found to be effective to classify DR senses \cite{zhao-etal-2023-infusing,chan-etal-2023-discoprompt,jiang-etal-2023-global}. 

\subsection{Natural language prompts for zero/few-shot inferences}

LLMs have demonstrated impressive performance in reasoning tasks with/without in-context examples.  This has fostered extensive research on effective ways to design natural-language prompts to query LLMs. For example, it was found that prompting the LLMs to provide explanations is useful \cite{reynolds2021prompt,lampinen-etal-2022-language}. In particular, the generation of intermediate reasoning steps before the final answer \citep[\textit{Chain-of-Thought},][]{wei2022chain} can significantly improve the performance, even without few-shot examples \cite{kojima2022large}. Subsequent works further investigated how the reasoning steps could be verified \cite{wang2022self,wang2023metacognitive} and decomposed \cite{yao2023tree,besta2023graph}. In particular,  \citet{dhuliawala2023chainofverification} found that verification questions are typically answered with higher accuracy than the original question. The current work therefore also investigates the effectiveness of formulating DR classification prompts directly as verification questions.

It is yet unclear if zero-shot prompting approaches can also be applied to abstract and subtle linguistic interpretations, which additionally require world knowledge, like discourse coherence. Related to our work, \citet{ostyakova-etal-2023-chatgpt} compare human and ChatGPT's annotations of dialogue functions.  They found that decomposing the 32-way classification task to a tree of binary questions largely improves the inference performance. However, the development of the structural prompts involves heavy engineering, and the methodology was only evaluated on a small subset of $189$ utterances.
We investigate alternative methods to disassemble the established task of implicit DR recognition and evaluate the LLM's performance against large samples of expert-annotated and crowdsourced data.

\subsection{Zero / few-shot DR annotation }

To our knowledge, \citet{chan2023chatgpt} is the only work that has investigated the zero-shot performance of LLMs in DR recognition. They evaluated ChatGPT's performance in the classification of DRs in PDTB 2.0 and DiscoGeM using a multiple-choice template that lists the $11$ Level-2 and $18$ Level-3 DR label options of the two corpora respectively. It was found that most explicit DRs could be recognized correctly (F1 > $60\%$ for most DR types).  However, the performance for implicit DRs was much poorer. It achieved $16.20\%$ F1 and $24.54\%$ accuracy on 11-way Level-2 classification of the PDTB~2.0 Ji-test set and F1 < $15\%$ for most DR types, considerably lagging behind the SOTA supervised model \citep[$49.03\%$ F1 and $64.58\%$ accuracy,][]{chan-etal-2023-discoprompt}. Pairing the relation label options with typical DCs was shown to improve the performance while the few-shot performance varied depending on the in-context examples provided and could degrade the performance significantly.
We re-ran their prompts using GPT-4 for comparison.

In addition, the performance of prompt-based inference has been shown to differ strongly between different relation subsets.  Among the DRs defined in PDTB (see Table \ref{tab:label_dist}), causal and temporal relation reasoning are established as separate tasks with dedicated datasets, such as COPA \cite{gordon-etal-2012-semeval} for causal relations and TB-Dense \cite{cassidy-etal-2014-annotation} for temporal relations. The formats and designs of these tasks and datasets are not the same as the DR recognition tasks: typically, the task asks for more fine-grained causal or temporal relations, given that the relations are present in the text.
\citet{chan2023chatgpt} showed that ChatGPT outperforms supervised baselines in causal relation detection, but underperforms in temporal relation classification. \citet{gao2023chatgpt} instead reported that ChatGPT can be biased to over-predict causality, depending on the prompting format, and can only capture explicit causality.

\section{Methodology}
Prompting LLMs to classify among specific labels typically requires listing all valid options.  The input becomes even longer when an example per class is included for in-context learning.  Instead, we propose several methods to break down the 14-way DR classification task into smaller sub-tasks, which are described in details below.

\subsection{Two-step DC insertion prompt}

 This approach adapts the two-step method used to crowdsource DR annotations \citet{yung-etal-2019-crowdsourcing} into a two-step prompt.  In the first step, the LLM is prompted to generate a word or phrase that represents the relation between two given arguments.  As a continuous conversation, a second prompt asks for a forced choice among a subset of options.  The provided options are based on the free insertion in the first step, following the mapping used in the original crowdsourcing method.  For example, \textit{however} could be generated in the first task, but it can ambiguously signal a \textsc{contrast} or \textsc{concession} relation. The second step, which is a forced choice between \textit{despite} and \textit{in contrast}, serves as a verification question to identify fine-grained DRs.
 An example of the input and output is shown in Figure \ref{fig:2step_prompt} in Appendix \ref{sec:example_appendix}.

 This method assumes that DRs can be interpreted and produced through the lexical semantics of DCs and does not require specific training about the definitions of the DRs.  As in the original crowdsourcing method, we did not include in-context examples in the prompt\footnote{This decision is also because we found that, as reported in previous work \cite{chan-etal-2023-discoprompt}, the LLM's prediction varies a lot depending on the examples provided in the prompt, adding more uncertainty to the effectiveness of the prompt.}.

\subsection{Per-class binary prompt}

This method decomposes the multi-way DR classification task into independent binary prompts, e.g., \textit{"does the discourse relation between the provided arguments represent a \textsc{asynchronous} relation?"}.  One binary question is used for each class, so 14 prompts are necessary for each instance of DR (for the 14 Level-2 DR sense defined in PDTB 3.0). A short description of the relation type, taken from the annotation manual \cite{prasad2007}, is also included (see Figure \ref{fig:binary_prompt} for an example). For each binary question, one positive and one negative example, also taken from the annotation manual, of the particular relation are provided in the prompt: the positive example is the demonstrative example of the relation and the negative example is the demonstrative example of another relation type that has a different top-level sense category. 

This method can produce multiple labels because the GPT model can answer \textit{yes} to several of the binary prompts.  This is particularly relevant to DR inference because multiple DRs can co-occur and simultaneously be interpreted by different reasoning traits \cite{scholman2017specification}.  In the crowdsourced DiscoGeM corpus, 
most relations are annotated with two or more DR senses (see Table \ref{tab:F1_pdtb}),
against which the multiple predicted labels can be compared. 

It is nonetheless necessary to combine the answers of the binary prompts into a single DR label in order to compare with the single gold labels in PDTB. We use the multiple-choice (MC) prompt \cite{chan2023chatgpt} that lists all DR options that were answered with \textit{"yes"} in the binary questions and ask for the best choice among the given options. In case all DR senses were answered with \textit{"no"}, all the options are provided in the MC step. The input and output of all the binary questions are included in the context. An example of the input and output is shown in Figure \ref{fig:binary_prompt} in Appendix \ref{sec:example_appendix}.  We also tried asking for a confidence score for the answer to the binary questions, as documented in Figure \ref{fig:binary_prompt}. However, since nearly all answers were assigned the same confidence score, we ignored these scores in the subsequent analyses. 

\subsection{Per-class verification prompt} \label{sec:verification}
This method also breaks down the multi-way classification task into individual per-class prompts, but instead of a straightforward yes-no question, we formulate the binary question as a verification question. To do so, we make use of the hierarchical nature of the DR senses.  For example, to classify whether an instance is a \textsc{asynchronous} relation, we ask \textit{"which argument (Arg1 or Arg2) describes an event that precedes the other? Options: Arg1, Arg2, None"}, where the answers Arg1 and Arg2 correspond to the \textsc{asynchronous} sub-classes \textsc{succession} and \textsc{precedence} respectively\footnote{For the non-directional senses such as \textsc{conjunction} and \textsc{synchronous}, we derived verification questions based on finer-grained definitions of these senses, e.g., \textit{are the situations in Arg1 and Arg2 completely, partially or not overlapped in terms of time?}}. The instance is classified as a \textsc{asynchronous} relation if either \textsc{Arg1} or \textsc{Arg2} is generated.  

In other words, the answer to the verification question provides an explanation to justify the sense of the DR without stating the label, e.g. \textit{Arg1 describes an event that precedes Arg2, (that's why the relation between Arg1 and Arg2 is \textsc{causal}).}

Similarly, one positive and one negative example, in the form of the verification questions, are included in each binary prompt and a multiple-choice prompt is used to choose the best option from the multiple positive answers.  An example of the per-class verification prompt is shown in Figure \ref{fig:verification_prompt} in Appendix \ref{sec:example_appendix}.

\section{Experiment}

We conduct our experiment using the state-of-the-art version GPT model from OpenAI gpt-4 (queried in December 2023).  The experiments are implemented using the API provided by OpenAI. We evaluate the results against the annotations in PDTB 3.0 and DiscoGeM.  

\subsection{Data}

\paragraph{PDTB 3.0} is the largest discourse-annotated resource in English.  The texts are news articles from the Wall Street Jounals. We evaluate our method to classify 14 Level-2 relation types with more than 10 instances in the sections 21 and 22 of the PDTB3.0 (i.e. the Ji-testset), following the setup of previous works \cite{kim-etal-2020-implicit,xu2023dual}. Most items in the PDTB 3.0 are labelled with a single DR labels but a number of relations are annotated with two labels. 

\paragraph{DiscoGeM 1.0} is a crowdsourced discourse resource in English that includes texts from multiple genres: European Parliament preceedings, Wikipedia articles, and literature.  Each implicit DR in the corpus was labelled by $10$ crowdworkers, following the sense definitions of PDTB 3.0 and using the two-step DC insertion method \cite{yung-etal-2019-crowdsourcing}.  We evaluate our method to classify Level-2 relation types with over 10 instances in the test set of the corpus,\footnote{The count is based on the single majority label. The included 7 Level-2 relation types are also the most frequent relation types in the whole corpus.} excluding instances with the majority label \textsc{differentcon}, which means the DR sense is undetermined.

The predicted DR sense of each instance is compared against 1) the single majority label, which is the label that has the most votes. In case of a tie, one of the most voted labels are randomly selected; and 2) the multiple majority labels, which is the set of labels that have two or more votes. If none of the labels have two or more votes, the single majority label is used.
The distributions of the labels in both test sets are shown in Table \ref{tab:label_dist}. 

\begin{table}[h]
    \centering \small
    \begin{tabular}{@{}l|lll}
        \hline
        \textbf{Level-1.Level-2 labels} & \textbf{PDTB} & \textbf{DG sing} & \textbf{multi.} \\
        \hline
        Comparison.Concession &96 &77 & 16 \\
        Comparison.Contrast& 53& 26 & 6\\
        Contingency.Cause &384 & 402 & 116\\
        Contingency.Cause+Belief& 14 & - & -\\
        Contingency.Condition &14& - & -\\
        Contingency.Purpose &59& - & -\\
        Expansion.Conjunction &236& 382 & 125\\
        Expansion.Equivalence& 30& - & - \\
        Expansion.Instantiation& 123& 58 & 5\\
        Expansion.Level-of-detail& 208& 207 & 48\\
        Expansion.Manner &17& - & -\\
        Expansion.Substitution& 25& - & -\\
        Temporal.Asynchronous &102& 100 & 27\\
        Temporal.Synchronous& 35 & - & -\\
        2 labels & 67  & 0 & 589\\
        3 labels & 0 & 0 & 282 \\
        4 labels & 0 & 0 & 38 \\
        \hline
        \textbf{Total} & 1463 & 1252 & 1252 \\
        \hline
    \end{tabular}
    \caption{Distribution of the level-2 labels in the PDTB~3.0 Ji testset and the DiscoGeM~1.0 testset}
    \label{tab:label_dist}
\end{table}

\subsection{Baseline}
\citet{chan2023chatgpt} evaluated the MC prompt on $11$ PDTB 2.0 Level-2 relations and $18$ DiscoGeM Level-3 relations, using the \textit{gpt-3.5-turbo} model of ChatGPT.  We reran this standard prompt to classify $14$ PDTB 3.0 Level-2 senses and $7$ DiscoGem Level-3 senses, using \textit{gpt-4}.  Specifically, we use the classification prompt where each DR option is paired with a typical DC. Since the performance with in-context examples was found to be unstable and would require extra long inputs, we did not include examples in this implementation.  We modified the options from the $11$ PDTB 2.0 Level-2 labels to the $14$ PDTB 3.0 labels and refined the DCs attached to the DR options, by including DCs for both sub-types of Level-2 labels (e.g., \textit{before / after} for a \textsc{asynchronous} relation); or using less ambiguous connectives (e.g., \textit{in contrast} instead of \textit {however} for \textsc{concession}). The prompt template is shown in Figure \ref{fig:MC_prompt} in Appendix \ref{sec:example_appendix}.

Following \citet{chan2023chatgpt} and other previous works on supervised implicit DR classification, we prompt the LLM to generate DR labels given the two identified arguments according to the original corpus annotation. The retrieval of implicit DR arguments in DiscoGeM is trivial, because they are defined as two consecutive sentences that are not connected by an explicit DC. PDTB 3.0, however, also includes \textit{intrasential} implicit DRs and the identification of these DRs and their arguments require another annotation step. Before validating the possibility of a fully automated discourse annotation pipeline, we focus on implicit DR annotation under a simplified setup.

\subsection {Results}

\subsubsection{Evaluation on single-sense DRs}

First, we look at the comparison of different methods evaluated against PDTB 3.0, which is shown in Table \ref{tab:results_pdtb}. 
In addition to the baseline MC prompt, we also compare the results with three generic baselines: random, always \textsc{conjunction}, and always \textsc{cause}. The latter two DR types are the most common categories of implicit DRs in both corpora.  In addition, the performance of state-of-the-art supervised models for implicit DR classification are listed as a reference.

\begin{table*}[h]
    \centering \small
    \begin{tabular}{@{}l|ll|ll|ll@{}}
        \hline 
       &&\multicolumn{4}{c}{PDTB 3.0 (Ji-test)} \\
       &\textbf{per-item}&\textbf{avg. input}& \multicolumn{2}{c|}{Level-1 4-way} &
       \multicolumn{2}{c}{Level-2 14-way}\\
       \textbf{Models} & 
       \textbf{prompts} &
       \textbf{tokens} &
       \textbf{macro F1} & \textbf{Acc.} &
       \textbf{macro F1} & \textbf{Acc.}  \\
        \hline
        \textit{supervised models} &&&&&\\
        GOLF$_{large}$ \citep{jiang-etal-2023-global} &-& - &
        $74.21$ & $76.39$ & $60.11$ & $66.42$\\
        PEMI \citep{zhao-etal-2023-infusing} &-&  - &
        $69.06$ & $73.27$ & $52.73$ & $63.09$ \\
        CP-KD$_{large}$ \cite{wu-etal-2023-connective} &-&  - &
        $75.52$ & $78.56$ & $52.16$ & $ 67.84$  \\
        \hline
         \textit{baseline} &&&&&\\
         random &-& - &$24.24$&$33.08$&$6.34$&$7.66$\\
         all \textsc{conjunction} &-& - &$15.75$&$47.44$&$1.97$&$16.20$\\
         all \textsc{cause} &-& - &$12.68$&$35.89$&$3.04$&$27.75$\\
         \hline
        1) 14-way MC \cite{chan2023chatgpt}  &$1$& $245$ &$45.80$&$50.03$&$26.12$&$36.84$\\ 
        2) two-step DC insertion & $2$& $99$ &$23.44$&$30.49$&$6.02$&$15.52$\\
        3) per-class binary (avg. $3.62$ labels)  & $14$ &$2597$&- &($61.52$)&-&($53.79$)\\
        \> + multi-way MC &$1$& $120$ &$41.76$&$47.16$&$19.66$&$30.69$\\
        4) per-class verification  (avg. $7.67$ labels)  & $14$ & $3873$ &- &($95.56$)&-&($89.33$)\\
        \> + multi-way MC &$1$& $167$ &$47.53$&$52.84$&$25.77$&$36.98$\\
        \hline
    \end{tabular}
    \caption{Results of the PDTB 3.0 Ji-testset.  
    The average input token counts are calculated using the BPE tokenizer provided by OpenAi\footnotemark.
    The 4-way Level-1 evaluation is calculated by mapping the Level-2 predictions to Level-1 based on the sense hierarchy.
    To calculate the accuracy, a prediction is counted as correct if it matches one of the gold labels.
    Values in brackets refer to soft-match scores: any overlap between the predicted multiple labels and the gold labels is counted as correct.
    }
    \label{tab:results_pdtb}
\end{table*}

The following can be observed from Table \ref{tab:results_pdtb}: 
\begin{enumerate}
    \item inference by \textit{gpt-4} achieves only about half of the performance of the supervised models ($36.84\%$ vs.~$67.84\%$ on Level-2 accuracy); 
    \item the two-step DC insertion prompt performs poorly, achieving less than half of the performance of other prompts ($15.52\%$);
    \item  the per-class verification + MC aggregation method performs similarly with the baseline 14-way MC prompt ($36.98\%$ vs $36.84\%$), while the per-class binary method significantly underperforms ($30.69\%$).
\end{enumerate}

\footnotetext{\url{https://github.com/openai/tiktoken}}

\citet{chan2023chatgpt} reported that the per-class predictions by the standard multi-way MC prompt achieved an accuracy of $20.31\%$ and F1 of $10.73\%$ for the 11-way classification of Level-2 senses in PDTB 2.0, using \textit{gpt-3.5-turbo}. For the 14-way classification of the PDTB 3.0 senses, using \textit{gpt-4}, the accuracy and F1 are $36.84$ and $26.12$ respectively, which have considerably improved but are still far from a level of satisfactory reliability.
\begin{table}[htpb]
\small \centering
\begin{tabular}{l|lll|lll}
    \hline
    & \multicolumn{3}{c}{14-way MC} & \multicolumn{3}{c}{per-class vf.}\\
\textbf{labels} & \textbf{P} & \textbf{R} & \textbf{F1} 
 & \textbf{P} & \textbf{R} & \textbf{F1} \\
    \hline
Conjunction &	.52	&.27	&.36 & 	.50&	.49	&.49\\
Cause&	.49&	.45	&.47&.49&	.37	&.42\\
Cause+Belief&	.10&	.07&	.08&.00&	.00	&.00\\
Condition	&.02	&.07&	.03&.02	&.07&	.03\\
Purpose&	.55	&.90&	.68&.50&	.79&	.61\\
Contrast	&.14&	.54&	.22&.14&	.42&	.21\\
Concession	&.17&	.01&	.02&.14&	.03&	.05\\
Asynchronous&	.23&	.60	&.33&.26	&.58&	.36\\
Synchronous	&.14	&.28&	.19&.12	&.21	&.15\\
Level-of-detail&	.56	&.11&	.18&	.48&	.10&	.17\\
Instantiation	&.50	&.51	&.51&.44	&.51&	.47\\
Equivalence&	.55	&.20	&.29&.24&	.33	&.28\\
Manner&	.00&	.00	&.00&.00	&.00	&.00\\
Substitution	&.27&	.32	&.29 &.45&	.30&	.36\\
\hline
macro F1 &&& .26 &&&.26 \\
\hline
\end{tabular}
\caption{F1 scores on PDTB 3.0 Level-2 label prediction with the 14-way MC prompt and the per-class verification + MC aggregation method}
    \label{tab:F1_pdtb}
\end{table}

\begin{figure*}[htpb]
    \centering
    \includegraphics[width=0.495\textwidth]{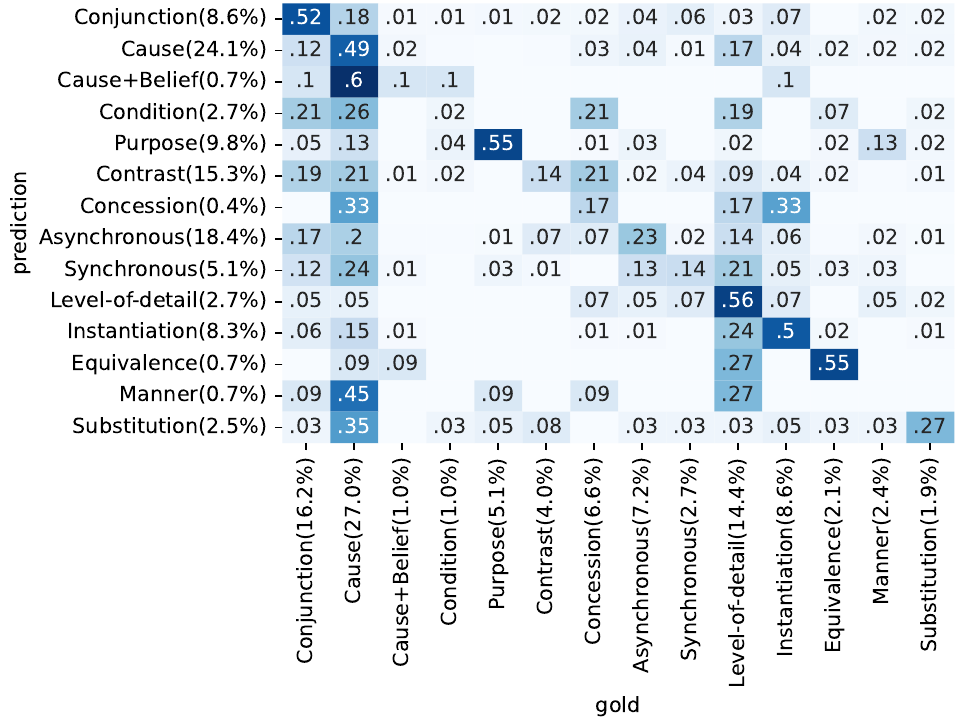}
     \includegraphics[width=0.495\textwidth]{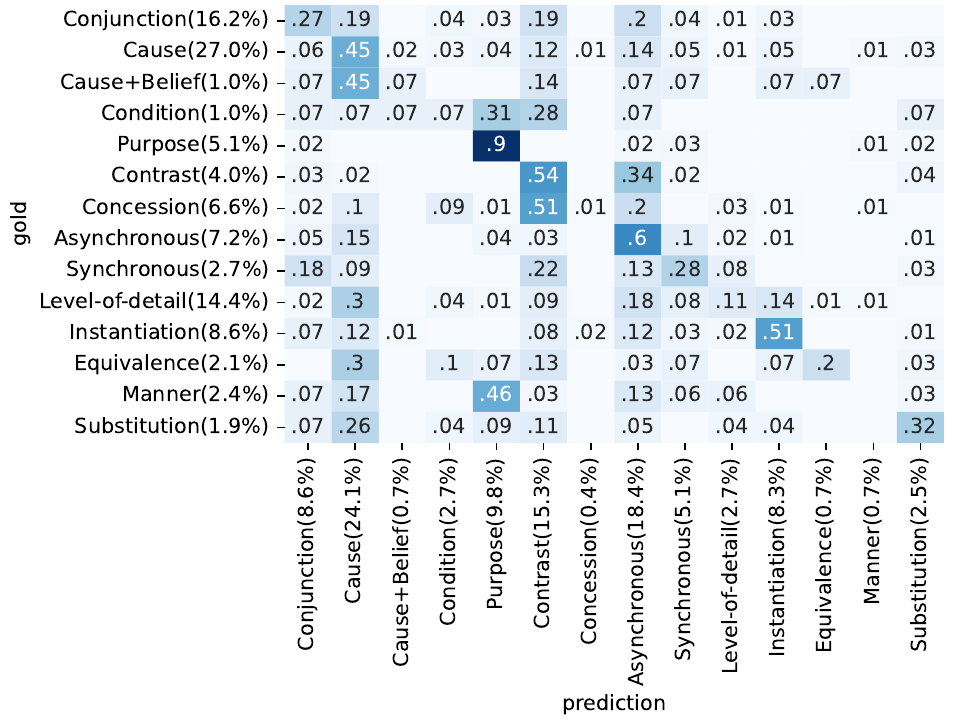}
     
    \caption{Confusion matrices comparing the gold and predicted labels in the \textbf{PDTB 3.0 test set} using the \textbf{MC prompt}. The distribution in the left figure is normalized by the predicted class, i.e. the diagonal corresponds to the \textbf{precision}; while the distribution on the right is normalized by the gold class, i.e. the diagonal corresponds to the \textbf{recall}. The percentages in brackets are the overall distributions of the predicted and gold labels respectively. }
    \label{fig:cm_chan_pdtb}
\end{figure*}

Table \ref{tab:F1_pdtb} compares the per-class precision, recall and F1 scores between the 14-way MC prompt and the per-class verification + MC aggregation prompt, and Figure \ref{fig:cm_chan_pdtb} shows the corresponding confusion matrices of the predicted and gold labels in the PDTB 3.0 test set items.
It can be observed that the precisions are generally higher than the recalls, but the performances are drastically different among different classes, ranging from $0\%$ F1 for \textsc{manner} to $61-68\%$ F1 for \textsc{purpose}.  The per-class performance of the two methods is similar.  The main difference is the better performance of the MC method on \textsc{cause} and of the verification method on \textsc{conjunction}.  

Some confusion patterns are similar to those of humans \cite{robaldo2014corpus,sanders1992,scholman2017specification}. For example, the confusion between \textsc{contrast} and \textsc{concession} and \textsc{cause} and \textsc{cause+belief} -- \textsc{concession} and \textsc{cause+belief} were hardly predicted at all. 
In general, there is particular confusion with \textsc{cause} and \textsc{level-of-detail} relations (darker column on these two relations in the left matrix, Figure \ref{fig:cm_chan_pdtb}).
Specific verification prompts to tease apart these easily confused relations could potentially improve the performance.
%

The poor performance of the two-step DC insertion method suggests that ChatGPT cannot infer DRs in a fully lexicalized manner based on DCs similar to humans. It is necessary to explicitly specify the link between the DCs and the DR labels, as in the MC prompts.

The underperformance of the per-class binary prompt is due to too many labels being rejected in the binary question step. The soft-match accuracy of $53.79$ means that in nearly half of the questions, ChatGPT answered \textit{"no"} to the correct relation in the first step. In fact, in about one-tenth of the questions in the PDTB data, all relation senses received the output \textit{"no"}.  This suggests that it is necessary to adjust the threshold of detecting a particular relation sense, which is not trivial to prompt.

\begin{table*}[htpb]
    \centering \small \tabcolsep=5pt
    \begin{tabular}{@{}l|ll|ll|ll|ll@{}}
        \hline 
       &&\multicolumn{4}{c}{DiscoGeM testset} \\
       && 
       \textbf{avg.}&
       \multicolumn{2}{c|}{Level-1 single} &
       \multicolumn{2}{c|}{Level-2 single} &
       \multicolumn{2}{c}{Level-2 multi}\\
       &\textbf{per-item}& 
       \textbf{input}&
       \multicolumn{2}{c|}{4-way} &
       \multicolumn{2}{c|}{7-way} &
       \textbf{avg. per-}&\\
        \textbf{Models} & 
       \textbf{prompt} &
       \textbf{tokens} &
     \textbf{macro F1} & \textbf{Acc.} &
       \textbf{macro F1} & \textbf{Acc.} &
       \textbf{item F1} & \textbf{Acc.}  \\
        \hline
        \hline
         \textit{baseline} &&&&&&&\\
          random &-&-&$24.20$&$29.63$&$12.99$&$13.18$&$17.22$&($26.99$)\\
          all \textsc{conjunction} &-&-&$17.12$&$51.68$&$6.76$&$30.51$&$36.91$&($55.27$)\\
          all \textsc{cause} &-&-&$12.12$&$32.11$&$6.93$&$32.11$&$34.91$&($52.72$)\\
          \hline

         7-way MC \cite{chan2023chatgpt}  &$1$&$231$&$41.21$&$45.52$&$27.87$&$32.67$&$35.28$&($52.16$)\\

          per-class verif.  (avg. $3.87$ labels)  & $7$ &$2473$&-&($90.34)$&-&($80.51)$&$50.63$&($92.81)$\\
        \> + multi-way MC &$1$&$184$&$37.68$&$44.41$&$24.08$&$30.83$&$32.61$&($47.84$)\\
         \hline
    \end{tabular}
    \caption{Results of the DiscoGeM test set. The predicted Level-2 labels are evaluated against the single majority labels at two levels (Level-1 single and Level-2 single), and against the multiple majority labels at Level-2.   Values in brackets refer to soft-match scores: any overlap between the predicted multiple labels and the gold labels is counted as correct.  \textit{Average per-item F1} is the F1 score of the multiple predicted labels compared with the multiple gold labels of each item, averaged by the total number of items.}
    \label{tab:results_dg}
\end{table*}

\begin{figure*}[htpb]
    \centering
   \includegraphics[width=0.495\textwidth]{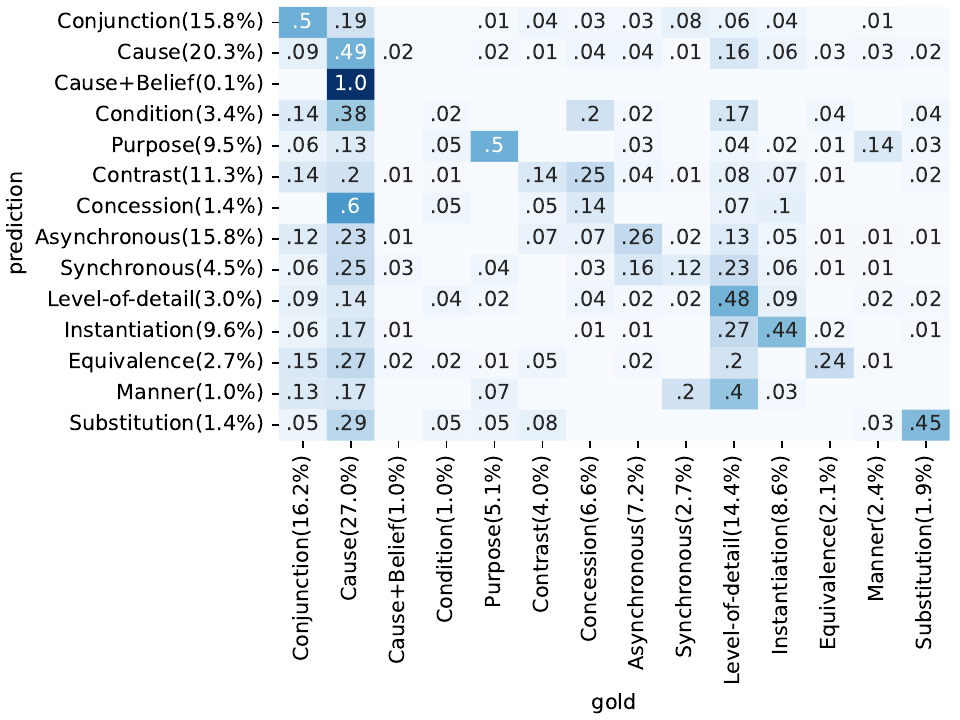}
     \includegraphics[width=0.495\textwidth] {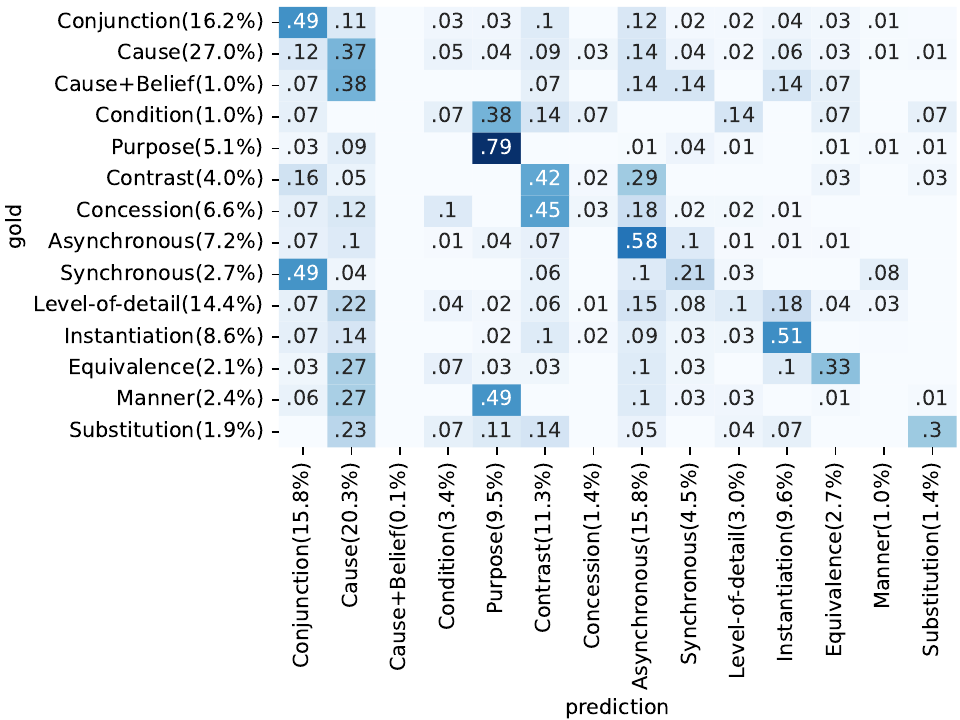}

    \caption{Confusion matrices comparing the gold and predicted labels in the \textbf{PDTB test set} using the \textbf{per-class verification prompt} with the MC aggregation step. }

    \label{fig:cm_verification_pdtb}
\end{figure*}

The performances of the MC prompt and the per-class verification prompt were found to be similar, as seen in Table \ref{tab:F1_pdtb} and Figure \ref{fig:cm_chan_pdtb} and \ref{fig:cm_verification_pdtb}. However, the MC prompt should be preferred since the cost of using the API, based on the number of prompts or the input tokens required for each item, is $15$ times less. While the F1 scores suggest that the MC-prompt is stronger in detecting \textsc{conjunction} and the per-class verification method is stronger in \textsc{cause}, the accuracies of both methods are too low to produce useful inferences for downstream tasks.

\subsubsection{Evaluation on multi-sense DRs}

One potential advantage of the per-class prompting methods over the MC prompt is the possibility of producing multiple labels by skipping the last MC step. It is not uncommon that several DR senses can be interpreted depending on the reader's perspective and multiple DR sense labels represent the semantics of the DR better.  

In Table \ref{tab:results_pdtb}, we see that the per-class verification method, without the multi-way MC step, reaches soft-match accuracy of $89.33\%$.  However, $7.67$ labels are predicted on average, and it is unclear how many of the predicted senses are actually valid. To further analyze the performance of ChatGPT's inference multi-sense DRs, we turn to the results of the crowdsourced DiscoGeM data. 
 \begin{table}[h]
 \small \centering
 \begin{tabular}{l|lll|lll}
     \hline
     & \multicolumn{3}{c}{7-way MC} & \multicolumn{3}{c}{per-class vf.}\\
 \textbf{labels} & \textbf{P} & \textbf{R} & \textbf{F1} 
 & \textbf{P} & \textbf{R} & \textbf{F1} \\
     \hline
 Conjunction  &   .53&  .29 & .38&  .56 & .35& .43\\
 Cause &  .58 & .30 & .39&.58 & .23 & .33\\
 Contrast    &    .05 & .62& .10& .09 & .54 &.15\\
 Concession   &   .24 & .10 & .14& .0  &   .0   &  .0\\
 Asynchronous  &  .31 & .76  &  .44& .21 &  .91  &  .33\\
 Level-of-detail &.32 & .33&.33& .26 & .10 & .15\\
 Instantiation  & .21 & .16 & .18& .20 & .59 & .29\\
 \hline
 macro F1 &&& .28 &&&. 24\\
 \hline
 \end{tabular}
 \caption{F1 scores on DiscoGeM Level-2 label prediction with the 14-way MC prompt and the per-class verification + MC aggregation method}
     \label{tab:F1_dg}
 \end{table}

Table \ref{tab:results_dg} compares the MC prompt and the per-class verification prompt with various baselines\footnote{Few supervised models have been evaluated on DiscoGeM and none of them are on a 7-way setting.}
and Table \ref{tab:F1_dg} compares the per-class F1s of the two prompting methods evaluated against the single gold sense label. In addition, we also evaluated the predictions against multiple gold sense labels.  In the DiscoGeM test set, each item has one to three labels (see Table \ref{tab:label_dist}). We calculate the per-item F1 score, which is the harmonic mean of precision and recall of the multiple predicted labels compared with multiple gold labels of each item\footnote{The average per-item F1 equals the accuracy if there are always one gold label and one predicted label. The macro F1 score, which is the arithmetic mean of all the per-class F1s, could not be calculated when most labels have multiple gold classes. In Table \ref{tab:F1_pdtb} and Figures \ref{fig:cm_chan_pdtb} and \ref{fig:cm_verification_pdtb}, items with two gold labels (< 5\% in the PDTB 3.0 test set) were treated as two separated items.}.

It can be observed in Table \ref{tab:results_dg} that: 
\begin{enumerate}
    \item the accuracy of ChatGPT's prediction is even lower in DiscoGeM, compared with PDTB 3.0 ($32.67\%$ v.s. $36.84\%$ with the MC prompt), and is similar to the \textit{all} \textsc{conjunction} and \textit{all} \textsc{cause} baselines;
    \item the per-class verification + MC aggregation method underperforms the 7-way MC prompt ($30.83\%$ v.s. $32.67\%$);
    \item the multiple predicted labels by the per-class verification without MC aggregation method considerably overlap with the multiple gold labels (average per-item F1 $50.63\%$).
\end{enumerate}

The lower accuracy on DiscoGeM can be attributed to the highly skewed label distribution, as seen in Table \ref{tab:label_dist}.  \textsc{cause} and \textsc{conjunction} each covers one-third of the relations in the data, such that the accuracy of these relations is highly reflected in overall accuracy despite similar per-class performance.  On the other hand, the lower accuracy of the results by the per-class verification is due to the lower F1 scores of the \textsc{level-of-detail} and \textsc{concession} in this dataset.

Evaluation against the single majority label of DiscoGeM involves a certain level of randomness since one of the majority labels is randomly selected when two or more labels have the maximum number of votes.  The multiple gold labels, on the other hand, are based on a threshold; they include all labels receiving $20\%$ or more votes among the 10 votes per item. We thus turn to the evaluation against the multiple Level-2 gold labels of DiscoGeM, which is shown in the right two columns of Table \ref{tab:results_dg}.

The soft-match accuracies, in brackets, are not directly comparable with the accuracies of single prediction against single gold labels because the chance agreement is higher.  Nonetheless, the soft-match accuracy of the multiple labels, which are $3.87$ labels on average, reaches $92.83\%$.  This means that in most cases the predicted labels overlap with the senses of the DRs.  The average per-item F1 is $50.63\%$, which is not too far from that between crowdsourced and expert multi-label annotations, which was found to be $58\%$ in a subset of the DiscoGeM corpus.

\section{Discussion and conclusion}
We set out to test ChatGPT's ability to infer implicit DR senses with the latest model and carefully engineered prompts.  Unfortunately, the low performance of implicit DR recognition could not be improved by sophisticated prompt engineering techniques that were successful in other tasks.  This points to the fact that either other prompting techniques are needed, or that implicit DR recognition simply cannot be solved under zero-shot or few-shot settings.  Knowledge acquired in other reasoning tasks does not seem to be transferrable to this task and supervised guidance to map the semantics of the arguments to the ambiguous and abstract DR labels is necessary.

We also performed smaller-scale experiments with other LLMs such as LLaMA \cite{touvron2023llama} but the performance was substantially worse even than \textit{gpt-3.5}. The training data of these other LLMs do not include PDTB nor DiscoGeM.
We found that ChatGPT is able to produce PDTB~2.0 labels even when the options are not provided in the prompt, suggesting that its training data should have at least included texts related to PDTB-style DR analysis (e.g., possibly an annotation manual or research article).  Therefore, strictly speaking, the inference made by ChatGPT is not completely zero-shot because it is informed about the DR labels. This may explain why the two-step DC insertion prompt, which does not involve any DR labels at all, totally failed in the task.

The underperformance of the per-class binary prompt suggests that prompting the discriminative comparison among all possible options at once is more accurate than separate detection of individual DR sense.  Too many relation senses were rejected when the model was presented with the binary choice of \textit{yes/no}; some of these rejected senses have been accepted when compared with an even more unlikely sense.  

The per-class approach, nevertheless, provides a framework to collect multi-label annotations, which is not only important to DR annotations but also to other tasks like natural language inference and sentiment analysis. We also experimented with running the MC prompts multiple times with a higher temperature setting, or explicitly asking for multiple labels in the prompt.  ChatGPT only occasionally produced multiple labels in these cases, possibly due to the dominance of single-label annotated data in its training history.

The better performance of the per-class verification approach compared with binary questions shows that the verification questions actually worked.  This approach is related to chain-of-thought prompting \cite{wei2022chain}; the identification of the arguments of the Level-3 sense justifies the presence of the Level-2 relation. We will experiment using this approach to refine the MC prompt. 

Another direction is to develop other approaches to disassemble the DR annotation task. Breaking down the multi-way classification task into smaller tasks was successful in dialogue structure annotation \cite{ostyakova-etal-2023-chatgpt}, using a heavily engineered step-by-step scheme (e.g. >~6 steps, each asking for specific features of the input).  Such a tailored annotation scheme might also be necessary to prompt implicit DR annotations.

\section{Limitations}
One of the limitations of the experiments is that we only queried the API once.  There could be variation in output between queries. In addition, the findings of the prompting techniques are limited to PDTB-styled DRs, and may not be generalized to other \merel{frameworks or} tasks. The experiment results are based on the specific templates we used. We did not implement nor compare other modifications, such as the choice of DCs in the options, which could potentially have an impact on the overall findings.

\section{Ethical consideration}
OpenAI's data collection complies with privacy laws\footnote{\url{https://help.openai.com/en/articles/7842364-how-chatgpt-and-our-language} \url{-models-are-developed}}. The PDTB 3.0 corpus is licensed under the LDC User Agreement. The text comes from the Wall Street Journal, which is publicly purchasable. DiscoGeM is publicly available on GitHub. The text comes from publicly available European parliament proceedings, Wikipedia articles, and novels. The annotation crowd-sourcing was approved by the Deutsche Gesellschaft für Sprachwissenschaft ethics committee. However, we did not check whether any of the raw texts contained any information that names or uniquely identifies individual people or offensive content, nor did we take any steps to anonymize it.

\section*{Acknowledgements}
This project is supported by the German Research Foundation (DFG) under Grant SFB 1102 (``Information Density and Linguistic Encoding", Project-ID 232722074).

\bibliography{custom}
\bibliographystyle{acl_natbib}

\appendix

\section{Appendix}\label{sec:cm_appendix}

\clearpage

 \begin{figure*}[htpb]
     \centering
      \includegraphics[width=0.495\textwidth]{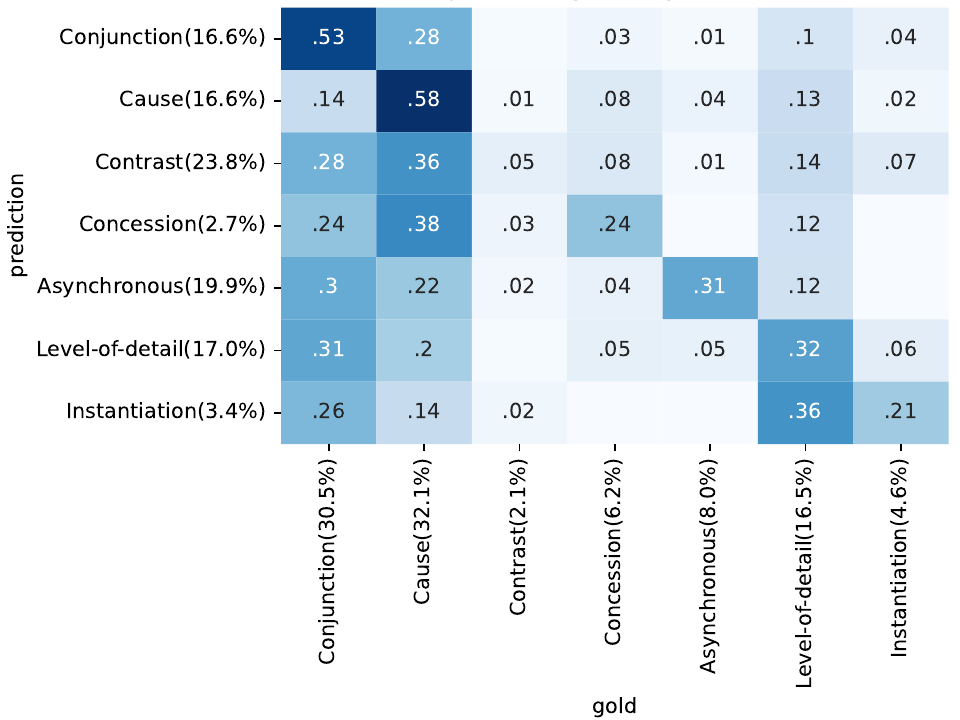}
     \includegraphics[width=0.495\textwidth]{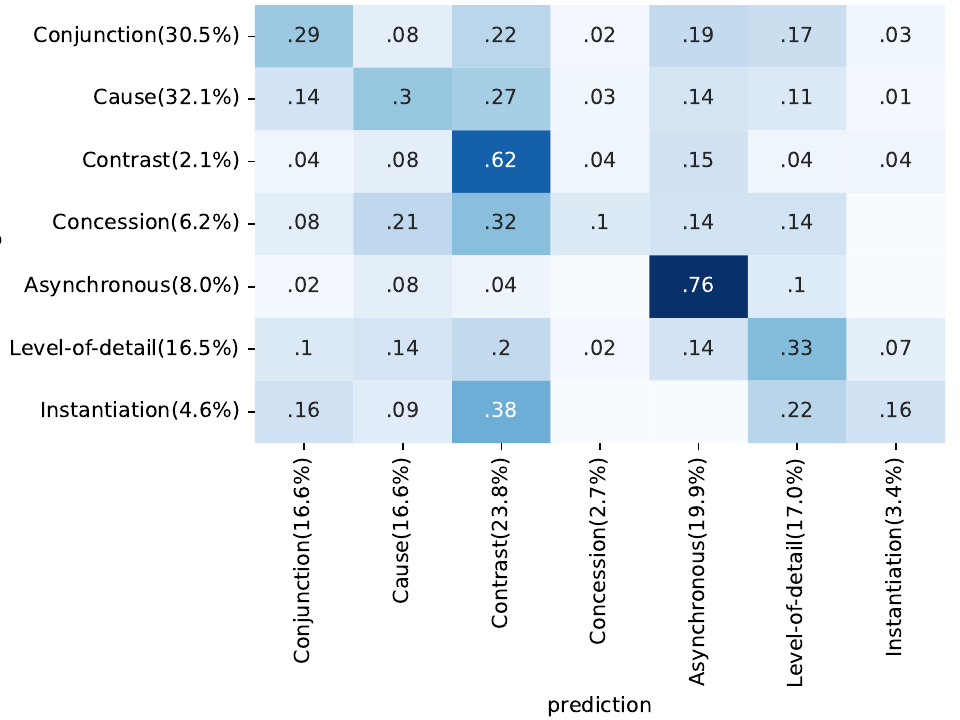}
     \centering
     \caption{Confusion matrices comparing the \textbf{single} gold and predicted labels in the \textbf{DiscoGeM} test set using the \textbf{MC prompt}.}
     \label{fig:cm_chan_dg}
 \end{figure*}

 \begin{figure*}[htpb]
     \centering
     \includegraphics[width=0.4950\textwidth]{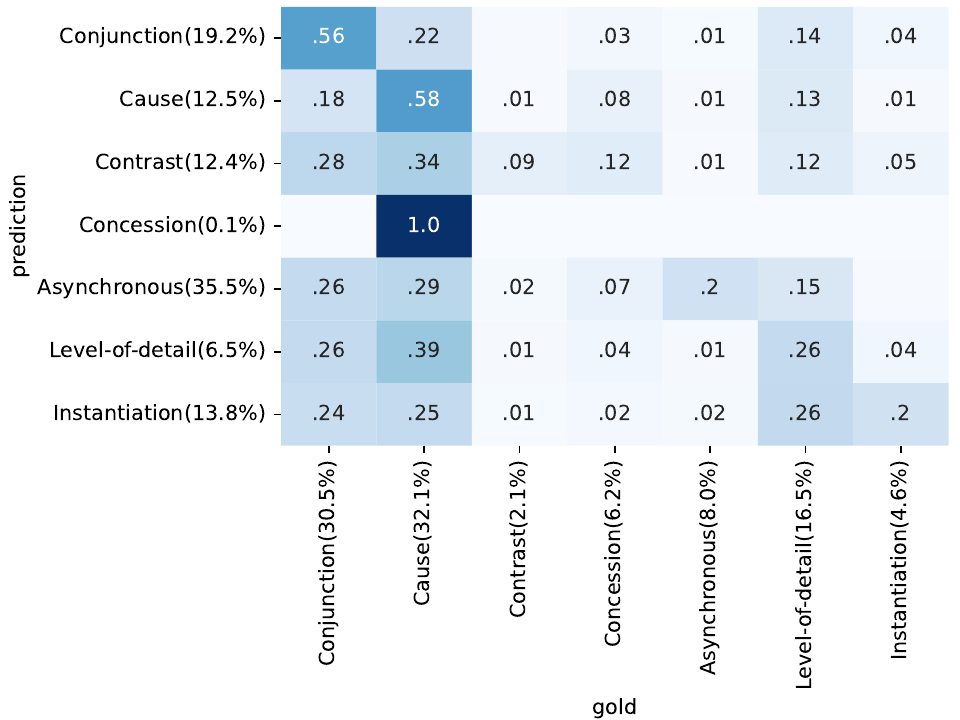}
      \includegraphics[width=0.495\textwidth]{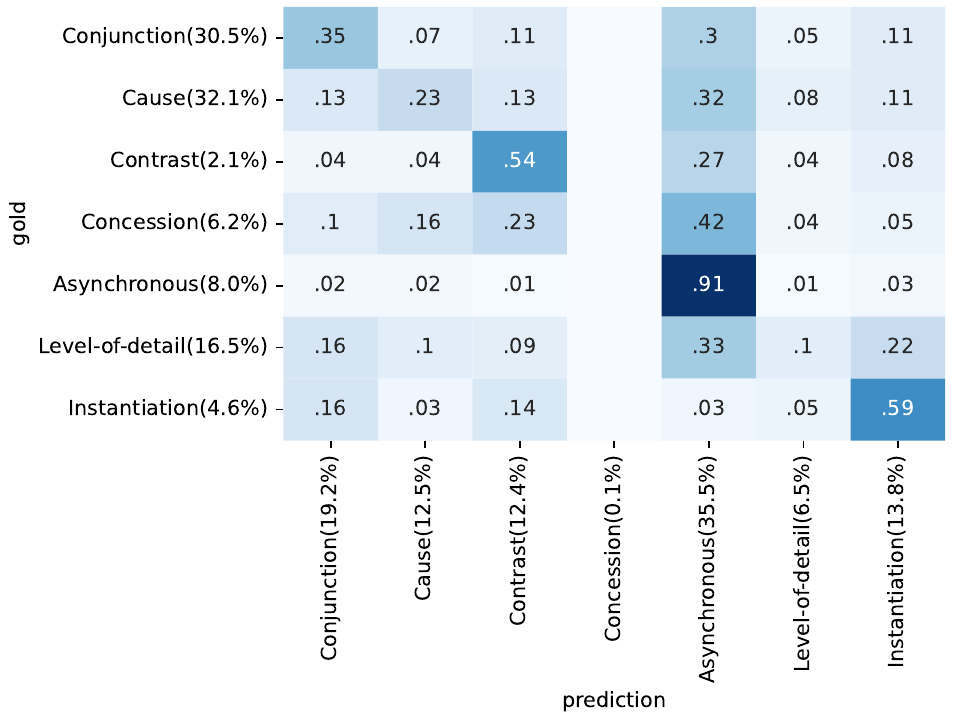}

     \caption{Confusion matrices comparing the \textbf{single} gold and predicted labels in the DiscoGeM test set using the \textbf{per-class verification prompt} with the MC aggregation step. }

     \label{fig:cm_verification_dg}
  \end{figure*}
\clearpage
\section{Appendix}\label{sec:example_appendix}
All prompts have \textit{"You are a language expert."} as the system content.
\begin{figure} [h]
\framedtext{\small
Task: Identify the most suitable option from the list below that describes the discourse relationship between the following pair of arguments.\\\\
Argument 1: We've got a product.\\
Argument 2: If you want it, you can get it.\\
\\
Options:\\
1. Temporal.Asynchronous, before / after\\
2. Temporal.Synchronous, at that time / while\\
3. Contingency.Cause, consequently / therefore\\
4. Contingency.Cause+Belief, considering this\\
5. Contingency.Condition, in that case / if \\
6. Contingency.Purpose, in order to / such that\\
7. Comparison.Contrast, on the contrary / in contrast\\
8. Comparison.Concession, despite this / even though\\
9. Expansion.Conjunction, in addition / also\\
10. Expansion.Instantiation, for example / for instance\\
11. Expansion.Equivalence, in other words\\
12. Expansion.Level-of-detail, specifically / in short\\
13. Expansion.Manner, how? / thereby\\
14. Expansion.Substitution, instead / rather\\
\\
Answer: ?
}
\caption{MC prompt adapted from \citet{chan2023chatgpt}}
    \label{fig:MC_prompt}
\end{figure}

\begin{figure} [h]
Free insertion step:
\framedtext{\small
Write down the connective word/phrase that best reflects the logical connection between these two arguments.\\\\ Argument 1: You build up a lot of tension. \\Argument 2: Working at a terminal all the day.
\\
\\Answer: ?
}

\bigskip
Forced-choice step:
\framedtext{\small
Select an option from the below list that best expresses the meaning of the phrase you have chosen in the first step.\\ \\Options:\\ 
1. in short\\ 2. for the reason that\\ 3. also\\
\\Answer: ?
}

\caption{Two-step prompt for implicit DR identification. Step 2's options are generated based on the free generation of Step 1.}
\label{fig:2step_prompt}
\end{figure}
\begin{figure} [h]

Binary step: one prompt is used for each DR class (i.e. 14 prompts per each item. Here is an example of the prompt for \textsc{Asynchronous}).
\framedtext{\small
Question: Does the discourse relationship between the provided arguments represent an Asynchronous relation?\\
\\
Description: Asynchronous relation describes a situation where one event is presented as preceding the other.\\
\\
Argument 1: The Artist sticks to a daily routine...\\
Argument 2: At night he returns to the condemned...\\
Answer: Yes\\
\\
Argument 1: The battle exceeds Justin's...\\
Argument 2: “I had no idea I was getting in so deep,” says...\\
Answer: No\\
\\
Argument 1: Capture the gaseous substance\\
Argument 2: And transport it to recycling center\\
Answer: ?\\\\
On a scale of 1-10,  1 being the lowest and 10 being the highest, Please express your confidence level in the prediction.
}

\bigskip
Multi-way MC step
\framedtext{\small
Task: Identify the most suitable option from the list below that describes the discourse relationship between the following pair of arguments.\\
\\
Argument 1: Capture the gaseous substance\\
Argument 2: And transport it to recycling center\\\\
Options:\\
1. Contingency.Cause, consequently / therefore\\
2. Expansion.Conjunction, in addition / also\\
3. Temporal.Synchronous, at that time / while\\
\\Answer: ?
}

\caption{Per-class binary prompt. Corresponding options (same as Figure \ref{fig:MC_prompt}) to the DRs answered with \textit{"yes"} in the binary step are listed as option in the MC step.}
\label{fig:binary_prompt}
\end{figure}

\begin{figure} [t]

Verification question step: one prompt is used for each DR class (i.e. 14 prompts per each item. Here is an example of the prompt for \textsc{Causal}.)

\framedtext{\small
Consider the discourse relation between Arg1 and Arg2, where Arg1 is "I trusted in his lordship's wisdom" and Arg2 is "I can't even say I made my own mistakes." Which argument (Arg1 or Arg2) gives the reason, explanation or justification of the effect described in the other argument?\\
Options: Arg1, Arg2, None\\
Answer: Arg1\\
\\
Consider the discourse relation between Arg1 and Arg2, where Arg1 is 'What is greatness?' and Arg2 is "What is dignity?" Which argument (Arg1 or Arg2) gives the reason, explanation or justification of the effect described in the other argument?\\
Options: Arg1, Arg2, None\\
Answer: None\\

Consider the discourse relation between Arg1 and Arg2, where Arg1 is 'The chain is reviewing its product list' and Arg2 is 'to avoid such problems' Which argument (Arg1 or Arg2) gives the reason, explanation or justification of the effect described in the other argument?\\
Options: Arg1, Arg2, None\\
Answer: ?
}

\bigskip
Multi-way MC step
\framedtext{\small
Task: Identify the most suitable option from the list below that describes the discourse relationship between the following pair of arguments.\\
\\
Argument 1: The chain is reviewing its product list\\
Argument 2: to avoid such problems\\\\
Options:\\
1. Comparison.Contrast, on the contrary / in contrast\\
2. Expansion.Conjunction, in addition / also\\
3. Contingency.Purpose, in order to / such that\\
\\Answer: ?
}

\caption{Per-class binary prompt. Corresponding options (same as Figure \ref{fig:MC_prompt}) to the DRs \textit{not} answered with \textit{"none"} in the verification question step are listed as options in the MC step.}
    \label{fig:verification_prompt}
\end{figure}

\end{document}